\documentclass[]{sjtu_sai}
\usepackage[toc,page,header]{appendix}
\usepackage{minitoc}
\usepackage{algorithm}
\usepackage{algpseudocode}
\usepackage{booktabs}
\usepackage{multirow}
\usepackage{siunitx}
\usepackage{makecell}
\usepackage{mathtools}
\usepackage{tabularx,array}
\usepackage[table]{xcolor}
\usepackage{graphicx}
\usepackage{wrapfig}
\usepackage{xspace}
\usepackage{amsmath}
\usepackage{amsfonts}
\usepackage{makecell}
\usepackage{amsmath}
\usepackage{wrapfig}
\usepackage[inline]{enumitem}
\usepackage{pifont}
\definecolor{myblue}{rgb}{0.21,0.49,0.74}
\definecolor{mygreen}{RGB}{0, 150, 80} 
\definecolor{mypurple}{RGB}{100, 50, 150} 
\newcommand{\cmark}{\ding{51}} 
\newcommand{\xmark}{\ding{55}} 
\definecolor{third}{RGB}{210, 230, 250} 
\definecolor{second}{RGB}{230, 220, 250} 
\definecolor{first}{RGB}{250, 225, 235} 
\DeclareMathAlphabet{\mathcal}{OMS}{cmsy}{m}{n}
\definecolor{NexoRed}{HTML}{B03A2E} 
\definecolor{SpaceBlue}{HTML}{1A5276}

\title{\textcolor{SpaceBlue}{Surround}\textcolor{NexoRed}{NEXO}: Ego-Centric Metric Bridging for Spatially Consistent Geometry in Autonomous Driving}

\author[1,2]{Shuai Yuan}
\author[1]{Runxi Tang}
\author[1]{Yuzhou Ji}
\author[1,2]{Fudong Ge}
\author[1,2]{Hanshi Wang}
\author[2]{Yifei Wang}
\author[2]{Xianming Zeng}
\author[2]{Jianyun Xu}
\author[2]{Xingliang Liu}
\author[1]{Yanfeng Wang}
\author[1,\dagger]{Zhipeng Zhang}

\affiliation[1]{School of Artificial Intelligence, Shanghai Jiao Tong University}
\affiliation[2]{Hello Inc.}
\contribution[\dagger]{Corresponding authors}

\abstract{Modern autonomous driving depends on accurate metric 3D understanding for perception, reconstruction, and planning, which in turn requires reliable multi-camera depth prediction. 
However, the outward-facing nature of vehicle-mounted surround-view camera rigs inherently limits visual overlap across views, challenging the correspondence-based assumptions that underpin conventional multi-view geometry. 
To bridge this gap, we present \textbf{SurroundNEXO}, named after the Spanish word \textit{nexo} for a geometric link, a low-overlap multi-camera metric depth framework that grounds cross-view reasoning in ego-centric geometry rather than dense visual correspondences. 
Instead of directly enforcing early global fusion, SurroundNEXO first assigns image tokens globally comparable ego-frame viewing directions through Ego-Ray Positional Encoding, then uses sparse LiDAR measurements as metric anchors to propagate absolute scale cues, and finally expands feature interaction progressively from view-local modeling to decomposed spatio-temporal reasoning and global integration. 
This design enables metric-scale depth prediction with improved spatial consistency across weakly overlapping cameras. 
Across low-overlap autonomous driving benchmarks, including  NuScenes, Waymo and DDAD, SurroundNEXO reduces single-view error by 33.2\%, improves cross-view consistency by 10.5\%, and enhances metric reconstruction quality by 25.6\% compared with SOTA methods. 
It further remains robust under extremely sparse depth prompts and exhibits strong zero-shot generalization to unseen camera layouts. }

\checkdata[Project Page]{\url{https://henryyuan429.github.io/papers/SurroundNEXO/}}
\begin{document}
\maketitle

\begin{figure}[t]
    \centering
    \vspace{-0.5em}
    \includegraphics[width=\textwidth]{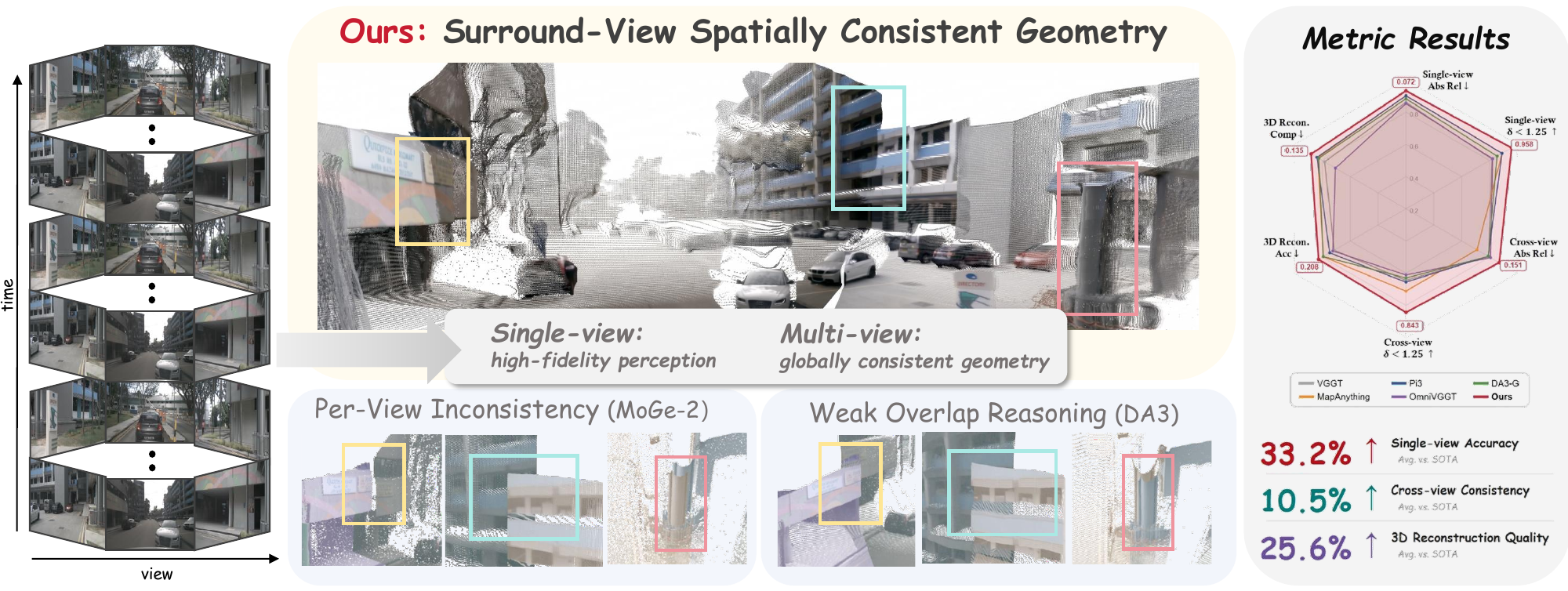}
    \caption{
\textbf{SurroundNEXO addresses metric depth estimation in low-overlap surround-view driving scenes.}
Instead of relying on dense visual correspondences, it links surrounding cameras through a shared ego-centric geometric reference and sparse metric cues, leading to improved single-view accuracy, cross-view consistency, and 3D reconstruction quality over state-of-the-art methods.
}
    \label{fig:teaser}
    \vspace{-0.5em}
\end{figure}

\section{Introduction}
\label{sec:intro}
Surround-view depth perception plays an indispensable role in modern autonomous driving, acting as the fundamental bridge between 2D visual inputs and 3D geometric reasoning. It underpins a wide spectrum of downstream applications, including 3D occupancy prediction\cite{zheng2023occworld,wei2023surroundocc,huang2023selfocc,huang2024gaussianformer,huang2024gaussianformer2,zuo2024gaussianworld}, bird's-eye-view (BEV) perception\cite{2022bevdepth,huang2021bevdet,li2022bevformer,li2022bevstereo,huang2022bevdet4d}, and high-fidelity scene reconstruction\cite{2025drivingforward,Lu2024DrivingRecon,wei2024omniscene,dggt,2025storm,shi2026unisplat}. Despite its importance, obtaining highly accurate and dense depth maps across all surrounding views remains notoriously difficult. Existing paradigms are fundamentally bottlenecked by the extreme sparsity of available LiDAR ground truth and the limited field-of-view (FoV) overlap between adjacent cameras. Consequently, models often struggle to propagate reliable geometric structures from the narrow overlapping regions to the vast non-overlapping areas, leading to severe depth degradation in complex driving scenarios.

To tackle these challenges, numerous efforts have been made, yet a critical dilemma remains between intra-view accuracy and inter-view consistency. While early self-supervised methods\cite{2022surrounddepth,vfdepth2022,r3d32023,2024m2depth,2024cvcdepth} fall short of practical performance requirements, recent monocular foundation models\cite{mogev22025,unidepthv22025,DepthPro2024,hu2024metric3dv2,metricanything2026} deliver impressive per-view accuracy. However, independent inference inevitably causes severe scale misalignment in overlapping regions due to the lack of explicit cross-view interaction (as shown in Fig.~\ref{fig:teaser}). To enforce global consistency, recent multi-view feed-forward models like VGGT~\cite{2025vggt} and DA3~\cite{depthanythingv3} introduce global attention mechanisms. Yet, deploying them in autonomous driving reveals a fundamental mismatch, as these mechanisms heavily rely on extensive co-visible regions to establish reliable correspondences. Given the extremely narrow shared regions in driving setups (\textit{e.g.}, $\sim$10\% in Waymo\cite{waymo}), these models often compromise per-pixel accuracy for consistency and fail to outperform standard monocular baselines. Even the latest driving-specific paradigm, DVGT\cite{2025dvgt}, struggles to resolve this dilemma. Its factorized, ``cross-shaped'' attention strictly confines information exchange to either isolated spatial views or temporal frames. This architectural constraint inherently hinders effective cross-view interaction across the narrow overlapping regions, leading to crucial information loss and failing to guarantee robust inter-view consistency.

These limitations prompt us to rethink a critical issue that \textit{How can we effectively bridge the gap between high-fidelity single-view perception and globally consistent 3D geometry without falling into the trap of architectural mismatch?} Delving into this issue, we identify that the limitations of current paradigms primarily stem from three representational and architectural bottlenecks directly tied to the aforementioned challenges.

\textbf{First} (\textbf{Alignment}), the severe scale misalignment in overlapping regions arises because independent per-view representations lack a unified 3D spatial awareness. To bridge this gap, we introduce Ego-Ray Positional Encoding (ERPE). By mapping patch-level image features to ray-aligned coordinates in the ego frame, ERPE equips each visual token with a view-aware yet globally consistent directional prior, effectively compensating for the lack of dense cross-view correspondences.
\textbf{Second} (\textbf{Scale}), even with unified spatial parameterization, feed-forward models inherently struggle to infer accurate metric scales due to the extreme sparsity of LiDAR supervision highlighted earlier. While previous depth injection methods\cite{2026mapanything} yield suboptimal accuracy, we propose a Sparse Metric Anchoring (SMA) module. SMA constructs reliable metric anchors directly from the available sparse LiDAR measurements, effectively propagating absolute scale cues from sparse physical observations to dense visual tokens.
\textbf{Third} (\textbf{Interaction}), the architectural mismatch in existing multi-view models forces a difficult choice between noisy global feature mixing across narrow overlaps and the overly restricted factorized attention seen in DVGT\cite{2025dvgt}. To navigate this exact trade-off, we design a Progressive Geometry Transformer (PGT). Rather than forcing premature cross-view fusion, PGT processes features in a coarse-to-fine manner. Early layers focus on stable single-view representation learning, intermediate layers introduce constrained cross-view interactions, and later layers perform comprehensive global integration. This progressive strategy ensures robust feature interaction without being compromised by the limited co-visible regions.

Building upon these insights, we propose SurroudNEXO, a novel surround-view metric depth framework tailored for low-overlap autonomous driving scenarios. Extensive experiments on standard autonomous driving benchmarks demonstrate its comprehensive superiority. Quantitatively, SurroundNEXO achieves a remarkable 33.2\% reduction in Absolute Relative Error (Abs Rel) and improves cross-view consistency metrics by 10.5\% compared to the state-of-the-art baseline\cite{depthanythingv3}. Furthermore, our method exhibits exceptional robustness to varying LiDAR sparsity patterns and significantly benefits downstream scene reconstruction tasks, boosting 3D reconstruction accuracy by 25.6\%.

In summary, our main contributions are threefold. $\spadesuit$ We propose SurroundNEXO, a novel feed-forward framework specifically designed to achieve metric-scale accurate and spatially consistent cross-view depth perception in low-overlap autonomous driving scenes. $\spadesuit$ We introduce ERPE to provide a shared spatial parameterization for view consistency, SMA to establish stable metric anchors for absolute depth accuracy, and PGT to facilitate highly efficient, coarse-to-fine feature interaction. 
$\spadesuit$ Extensive experiments demonstrate that our method achieves state-of-the-art performance across standard benchmarks, significantly improving single-frame accuracy and cross-view consistency.
\section{Related Works}
\label{sec:related}
\textbf{Multi-view Depth Estimation.}
Depth estimation has evolved from learning open-world relative geometry to recovering metric scale and, more recently, to incorporating external geometric cues as conditions. 
Previous work has progressively advanced depth estimation from generalizable monocular priors to metric prediction and sparse-conditioned inference. Early studies on relative\cite{MiDas2022,ViTforDense2021,depthanything2024,depth_anything_v22024,Marigold2024,DepthFM2025,Lotus2024,PPD2025} and metric monocular depth\cite{zoedepth2023,mogev22025,unidepth2024,unidepthv22025,DepthPro2024,yin2023metric3dv1,hu2024metric3dv2} established strong cross-domain priors and metric grounding strategies, while more recent work showed that sparse metric observations can further calibrate or densify monocular predictions\cite{2024promptda,priorda2025,2024marigolddc,2025omnicdc,lingbot-depth2026,metricanything2026,sun2026drivemvs}.
However, these methods remain largely image-wise or view-wise, they either recover depth only up to scale, depend heavily on learned camera priors, or use sparse cues mainly to refine a single view rather than to enforce consistency across a calibrated surround camera system.

This limitation is particularly critical in autonomous driving, where depth must be predicted jointly over multiple synchronized cameras. 
SurroundDepth~\cite{2022surrounddepth} and VFDepth~\cite{vfdepth2022} enhance self-supervised surround depth through cross-view feature interaction and volumetric feature fusion. 
R3D3~\cite{r3d32023} and M$^2$Depth~\cite{2024m2depth} introduce stronger spatial-temporal geometric reasoning via dense bundle adjustment or two-frame cost-volume fusion. 
More recent methods, including CVCDepth~\cite{2024cvcdepth} and Semi-SMD~\cite{xie2025semismd}, further improve cross-view consistency and metric supervision through explicit consistency constraints or semi-supervised fusion. 
However, surround-view modeling introduces a nontrivial trade-off between cross-view consistency and per-view depth fidelity. 
Therefore, a desirable model should preserve strong per-view visual priors while introducing lightweight geometric cues for metric scale and cross-view organization.

\textbf{Feed-forward 3D Geometry Models.}
Recent 3D geometry foundation models move beyond per-image depth prediction and instead directly infer scene geometry, camera parameters, and dense 3D structure in a feed-forward manner\cite{dust3r,mast3r}.
Subsequent models elevate this formulation from pairwise reasoning to native multiview inference. VGGT\cite{2025vggt} jointly predicts cameras, depth and point maps in a single transformer, while Pi3\cite{2025pi3} further removes the bias of reference-frame anchoring through permutation-equivariant relative geometry, and further to streaming memory for long sequences\cite{streamVGGT2025,2026infinitevggt}.
In another way, additional priors such as depth and poses are introduced as optional geometric conditions into a unified feed-forward reconstruction framework\cite{2026mapanything,omnivggt2025}. 
Recently, DA3\cite{depthanythingv3} extends this line with a unified any-view geometry model based on depth and ray prediction, representing one of the latest advances in feed-forward 3D geometry foundation models.
Since autonomous driving systems rely on critical camera calibration, the practical goal is not to discard such priors, but to use them in a generalizable manner.

More recently, this feed-forward geometry paradigm has been adapted to autonomous driving scenes. 
Driv3R~\cite{driv3r} learns dense 4D reconstruction from multi-view driving sequences by regressing per-frame point maps with sensor-wise memory and optimization-free multi-view alignment. 
DriveVGGT~\cite{2026drivevggt} introduces driving-specific constraints into VGGT-style geometry modeling, using temporal video attention and multi-camera consistency attention to handle calibrated camera rigs with limited overlap. 
DVGT~\cite{2025dvgt} further targets ego-centric metric 3D point-map reconstruction for driving scenes through spatial-temporal geometry transformers. 
\newpage
\section{Method}
\label{sec:method}
\begin{figure*}[!t]
    \centering
    \includegraphics[width=\textwidth]{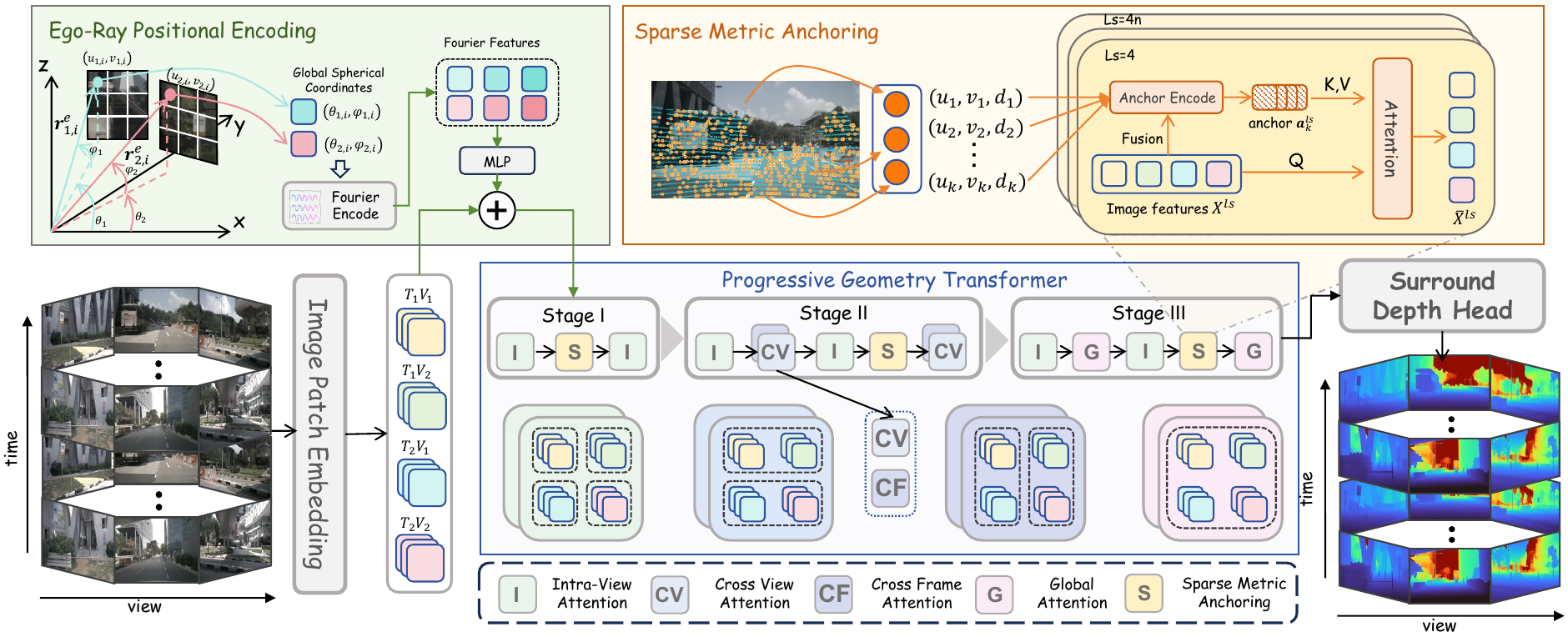}
    \caption{
\textbf{Overview of SurroundNEXO.}
Given surround-view images, camera rigs, and sparse LiDAR, SurroundNEXO grounds images with ego-frame ray priors and sparse metric anchors, then progressively expands feature interaction to produce metric and spatially consistent depth predictions.
}
    \label{fig:method}
\end{figure*}  

\subsection{Overall Architecture}
Fig.~\ref{fig:method} illustrates our spatially consistent depth estimation framework.
We argue that low-overlap multi-camera depth estimation requires three complementary conditions: a shared geometric reference for organizing tokens from different cameras, reliable metric cues for absolute scale recovery, and a progressive interaction strategy that avoids premature noisy cross-view fusion. 
Accordingly, we build a unified framework consisting three synergistic modules: 
\begin{enumerate*}[label=(\arabic*)]
    \item Ego-Ray Positional Encoding (ERPE), detailed in Sec.~\ref{sec:erpe}, maps each image patch to an ego-frame ray representation;
    \item Sparse Metric Anchoring (SMA), detailed in Sec.~\ref{sec:SMA}, injects sparse metric depth observations as anchor-based scale cues; and 
    \item Progressive Geometry Transformer (PGT), detailed in Sec.~\ref{sec:progressive}, gradually expands the feature interaction range from view-local modeling to cross-view and global integration.
\end{enumerate*}

\subsection{Ego-Ray Positional Encoding}
\label{sec:erpe}

In low-overlap surround-camera systems, direct visual overlap is often insufficient for establishing reliable token-level correspondences across views. 
We therefore introduce Ego-Ray Positional Encoding (ERPE), which assigns each image patch a globally comparable viewing direction in the ego frame. 
For the $i$-th patch in camera $v$, let $\tilde{\mathbf{p}}_{v,i}=[u_{v,i},v_{v,i},1]^\top$ denote the homogeneous coordinate of the patch center in the image plane, $K_v\in\mathbb{R}^{3\times3}$ denote the camera intrinsic matrix, and $R_{c_v\rightarrow e}\in\mathbb{R}^{3\times3}$ denote the rotation from the camera coordinate system to the ego coordinate system. 
We first compute the normalized camera-frame viewing ray and rotate it into the ego frame:
\begin{equation}
    \mathbf{r}_{v,i}^{e} 
    = 
    R_{c_v\rightarrow e} 
    \frac{K_v^{-1}\tilde{\mathbf{p}}_{v,i}}
    {\left\|K_v^{-1}\tilde{\mathbf{p}}_{v,i}\right\|_2}.
\end{equation}
Let $\mathbf{r}_{v,i}^{e}=(r_x,r_y,r_z)$. 
The ego-frame ray direction is parameterized by spherical angular coordinates:
\begin{equation}
    (\theta_{v,i},\phi_{v,i}) 
    = 
    \Pi_{\mathrm{sph}}(\mathbf{r}_{v,i}^{e}),
\end{equation}
where 
\begin{equation}
    \Pi_{\mathrm{sph}}(\mathbf{r})
=
\left(
\mathrm{atan2}(r_y,r_x),
\mathrm{atan2}(r_z,\sqrt{r_x^2+r_y^2})
\right).
\end{equation}
The corresponding normalized global spherical coordinates are
\begin{equation}
    u_{v,i}^{g}
=
\frac{\theta_{v,i}+\pi}{2\pi},
\qquad
v_{v,i}^{g}
=
\frac{1}{2}+\frac{\phi_{v,i}}{\pi}.
\end{equation}
Instead of using local image-plane coordinates, ERPE encodes the ego-frame angular coordinates with Fourier features:
\begin{equation}
    \mathbf{e}_{v,i}^{\mathrm{ray}}
=
\mathrm{MLP}_{\mathrm{ray}}
\left(
\gamma(\theta_{v,i},\phi_{v,i})
\right),
\end{equation}
where $\gamma(\cdot)$ denotes Fourier positional encoding, and $\mathrm{MLP}_{\mathrm{ray}}$ maps the encoded ray direction to $\mathbf{e}_{v,i}^{\mathrm{ray}}\in\mathbb{R}^{C}$, which has the same dimension as the visual token.
Given the patch token $\mathbf{x}_{v,i}^{0}\in\mathbb{R}^{C}$ after image patch embedding, we inject ERPE by additive conditioning:
\begin{equation}
    \tilde{\mathbf{x}}_{v,i}^{0}
    =
    \mathbf{x}_{v,i}^{0}
    +
    \mathbf{e}_{v,i}^{\mathrm{ray}}.
\end{equation}
This gives tokens from different cameras a shared ego-frame directional prior, allowing later interaction layers to reason about cross-view geometry without relying on dense visual overlap.

\subsection{Sparse Metric Anchoring}
\label{sec:SMA}
ERPE provides a shared spatial reference, but it does not determine the metric scale, which specifies the absolute depth in meters.
To inject metric information, we represent sparse depth observations as metric anchors and propagate their scale cues to dense image tokens through Sparse Metric Anchoring (SMA). 
For each camera view, we sample valid sparse depth points:
\begin{equation}
    \mathcal{P}_s =
\{(u_n,v_n,d_n)\}_{n=1}^{N},
\qquad d_n > 0,
\end{equation}
where $(u_n,v_n)$ is the normalized image coordinate and $d_m$ is the metric depth value. 
Then each sparse point is encoded as an anchor:
\begin{equation}
\mathbf{a}_{n}^{l}
=
\mathrm{MLP}_{\mathrm{fuse}}
\left(
\mathrm{Concat}
\left[
\phi_{\mathrm{pos}}(u_n,v_n),
\phi_{\mathrm{dep}}(d_n),
\phi_{\mathrm{vis}}^{l}
\left(
\mathrm{Bilinear}(X_v^l,u_n,v_n)
\right)
\right]
\right).
\end{equation}
where $\phi_{\mathrm{pos}}:\mathbb{R}^{2}\rightarrow\mathbb{R}^{C/4}$ encodes the normalized anchor coordinate with Fourier features, $\phi_{\mathrm{dep}}:\mathbb{R}^{1}\rightarrow\mathbb{R}^{C/2}$ encodes the metric depth, $\phi_{\mathrm{vis}}^{l}:\mathbb{R}^{C}\rightarrow\mathbb{R}^{C/4}$ projects the layer-$l$ visual feature sampled at the anchor location and $\mathrm{MLP}_{\mathrm{fuse}}$ fuses these components into a unified anchor embedding.
Given image token $\mathbf{x}_{i}^{l}$ and anchor $\mathbf{a}_{n}^{l}$, SMA performs asymmetric cross-attention where dense RGB tokens query sparse metric anchors:
\begin{equation}
    \mathbf{q}_i = W_q \mathbf{x}_{i}^{l},
\qquad
\mathbf{k}_n = W_n\mathbf{a}_{n}^{l},
\qquad
\mathbf{v}_n = W_v\mathbf{a}_{n}^{l}.
\end{equation}
To encourage physically meaningful local propagation, we add a spatial distance decay bias between the image patch coordinate $\mathbf{p}_i=(u_i,v_i)$ and the anchor coordinate $\mathbf{p}_n=(u_n,v_n)$:
\begin{equation}
    \alpha_{in}
=
\mathrm{Softmax}_{n}
\left(
\frac{\mathbf{q}_i^\top \mathbf{k}_n}{\sqrt{C}}
-
\frac{\|\mathbf{p}_i-\mathbf{p}_n\|_2^2}{\sigma_l^2}
+
m_n
\right),
\end{equation}
where $m_n$ masks invalid anchors and $\sigma_l$ is a layer-dependent spatial decay factor. 
The SMA residual is then computed as:
\begin{equation}
    \Delta \mathbf{x}_{i}^{l}
=
W_o
\sum_{n=1}^{N}
\alpha_{in}\mathbf{v}_n.
\end{equation}
Since SMA is inserted before the main attention operator of selected transformer blocks, metric cues are injected into RGB tokens before they participate in local, cross-view, or global attention:
\begin{equation}
\bar{X}^{l}
=
X^{l}
+
\Delta_{\mathrm{SMA}}^{l}(X^l,S),
\end{equation}
\begin{equation}
X^{l+1}
=
\mathcal{F}_l(\bar{X}^{l}),
\end{equation}
where $S$ represents sparse metric anchors, $\mathcal{F}_l$ denotes the attention operator used at layer $l$.
This pre-attention injection allows sparse metric anchors to calibrate visual tokens before cross-view or global feature aggregation, rather than correcting the features only after they have already been mixed.

\subsection{Progressive Geometry Transformer}
\label{sec:progressive}
In low-overlap driving scenes, directly applying full global interaction over all camera and temporal tokens from shallow layers may introduce noisy associations, since early features are still dominated by local appearance and weak geometric evidence. 
We adopt a progressive architecture that expands the interaction scope from view-local to global scales, thereby controlling the spatial-temporal extent of block interactions.

Let $X^l \in \mathbb{R}^{B\times T\times V\times N\times C}$ denote the token tensor at layer $l$, where $T$ is the number of sampled frames, $V$ is the number of cameras, and $N$ is the number of tokens per image. 
Before entering the progressive transformer, each token is first equipped with ERPE:
\begin{equation}
X^0 = \mathrm{PatchEmbed}(I) + E_{\mathrm{ERPE}}(\{K_v\},\{R_{c_v\rightarrow e}\}).
\end{equation}
At selected layers, SMA injects metric cues before the interaction of the current transformer block:
\begin{equation}
    \bar{X}^{l}
=
\begin{cases}
X^{l}+\Delta_{\mathrm{SMA}}^{l}(X^{l},S), &l\in l_{\mathrm{SMA}}\\
X^{l}, & l\notin l_{\mathrm{SMA}}
\end{cases}.
\end{equation}
The updated features are then fed into the transformer block:
\begin{equation}
X^{l+1}
=
\mathcal{F}_{s(l)}^{l}(\bar{X}^{l}).
\end{equation}
In our implementation, SMA is applied at a fixed interval of four transformer blocks, forming the injection set
$l_{\mathrm{SMA}}=\{3,7,11,\ldots\}$.
The operator $\mathcal{F}_{s(l)}^{l}(\cdot)$ denotes the transformer block at layer $l$ under the corresponding interaction stage.
This ordering ensures that sparse metric anchors calibrate visual tokens before they participate in local, decomposed spatio-temporal, or global interaction at layer $l$, rather than correcting already aggregated features after attention.

In the local stage, each image is processed independently by intra-view attention, so the model preserves the strong single-view appearance and depth priors inherited from the foundation backbone. 
This is important because early features are not yet reliable enough for unrestricted cross-view mixing. 
In the following stages, intra-view attention is still interleaved with broader interactions, so the model does not abandon stable per-image updates when expanding the interaction range.
In the decomposed stage, the model separately introduces cross-view interaction among cameras at the same timestamp and cross-frame interaction along the same camera stream. 
This separates spatial and temporal relation modeling, reducing interference between low-overlap cross-camera associations and temporal motion cues. 
In the global stage, all sampled views and frames are integrated at a higher semantic level. 
By this point, ERPE has already provided a shared ego-frame directional reference and SMA has injected sparse metric scale cues, so global aggregation operates on geometry-aware and scale-aware tokens rather than raw visual features.

Overall, the architecture first grounds visual tokens in a shared ego-centric reference, then recalibrates them with sparse metric cues before selected transformer blocks, and finally propagates these geometry-aware representations from local image features to multi-view and temporal context through progressive feature exchange.

\subsection{Training Objective}

\textbf{Data Processing.}
Our training data are constructed from multiple autonomous driving datasets with calibrated surround cameras, including NuScenes\cite{nuscenes}, Waymo\cite{waymo}, KITTI\cite{KITTI2013}, DDAD\cite{ddad}, and OpenScene\cite{openscene2023}. 
These datasets are sampled with the weights of 0.7, 1.75, 0.4, 0.6, and 1.5, respectively, to balance data quality and dataset scale.
After filtering invalid or incomplete scenes, the actual training set is smaller than the raw scene collection. 

\textbf{Sparse Depth Input Augmentation.}
To improve robustness to depth prior availability, we augment only the sparse depth input used by SMA while keeping dense pseudo-depth supervision unchanged.
At each training iteration, the sparse input is randomly assigned one of five modes: unchanged, light sparsification, moderate sparsification, extreme sparsification, or fully removed.
The final configuration uses probabilities of 0.50, 0.25, 0.15, 0.05, and 0.05 for these modes, respectively.
In the sparsified modes, we retain progressively fewer valid sparse points per image, while the zero mode removes the sparse input completely.
This exposes the model to different regimes, reducing over-reliance on a fixed sparse-depth density and improving robustness under degraded or missing depth observations.

\textbf{Dynamic Multi-view Sampling.}
To expose the model to diverse surround-view layouts under a bounded memory budget, we enable dynamic sequence sampling during training.
For each sampled scene, the loader randomly selects camera views and temporal frames from a local segment rather than using a fixed camera-frame layout.
The final configuration samples at least two views and two frames when possible, limits each iteration to at most 16 images, and randomly varies the image aspect ratio from 1.5 to 3.3 before resizing to the training resolution.

\textbf{Training Loss.}
Our training objective jointly supervises metric depth accuracy, local depth structure, cross-view geometric consistency, and prediction confidence. Given predicted depth $\hat{D}$, pseudo metric depth $D^{*}$, valid mask $\Omega$, and confidence map $C$, we optimize
\begin{equation}
\mathcal{L}
=
w_{\mathrm{silog}}\mathcal{L}_{\mathrm{silog}}
+
w_{\mathrm{grad}}\mathcal{L}_{\mathrm{grad}}
+
w_{\mathrm{cv}}(t)\mathcal{L}_{\mathrm{cv}}
+
w_{\mathrm{conf}}\mathcal{L}_{\mathrm{conf}} .
\end{equation}

Here, $\mathcal{L}_{\mathrm{silog}}$ is a scale-aware logarithmic depth loss for dense metric supervision, $\mathcal{L}_{\mathrm{grad}}$ is a multi-scale log-depth gradient loss that preserves local geometric structures, $\mathcal{L}_{\mathrm{cv}}$ enforces cross-view consistency by reprojecting predicted depth between calibrated neighboring cameras, and $\mathcal{L}_{\mathrm{conf}}$ regularizes the confidence prediction following VGGT~\cite{2025vggt}. 

\textbf{Training Implements.}
We initialize the model from the pretrained DA3-G\cite{depthanythingv3} and train it for 30k steps on 16 NVIDIA H20 GPUs with 96GB memory, which takes approximately nine days. 
Optimization is performed with AdamW using a learning rate of $1\times10^{-4}$ for newly introduced parameters and $1\times10^{-6}$ for the pretrained backbone, with weight decay 0.01, $\beta=(0.9,0.999)$, a cosine schedule, and 8k warm-up steps. 
\section{Experiments}
\label{sec:experiment4}
\subsection{Experimental protocol}
\textbf{Datasets.}
We evaluate our method on a broad set of autonomous driving datasets, covering diverse sensor configurations, scene layouts, and driving conditions. Specifically, we use the validation splits of NuScenes\cite{nuscenes}, Waymo Open Dataset\cite{waymo}, DDAD\cite{ddad}, KITTI\cite{KITTI2013}, and OpenScene\cite{openscene2023} for quantitative evaluation. 
To evaluate zero-shot generalization, we additionally test on Argoverse, which is not used during training. This setting allows us to examine whether the proposed method can transfer to unseen sensor setups and scene distributions without dataset-specific fine-tuning.

\begin{figure*}[!b]
    \centering
    \includegraphics[width=\textwidth]{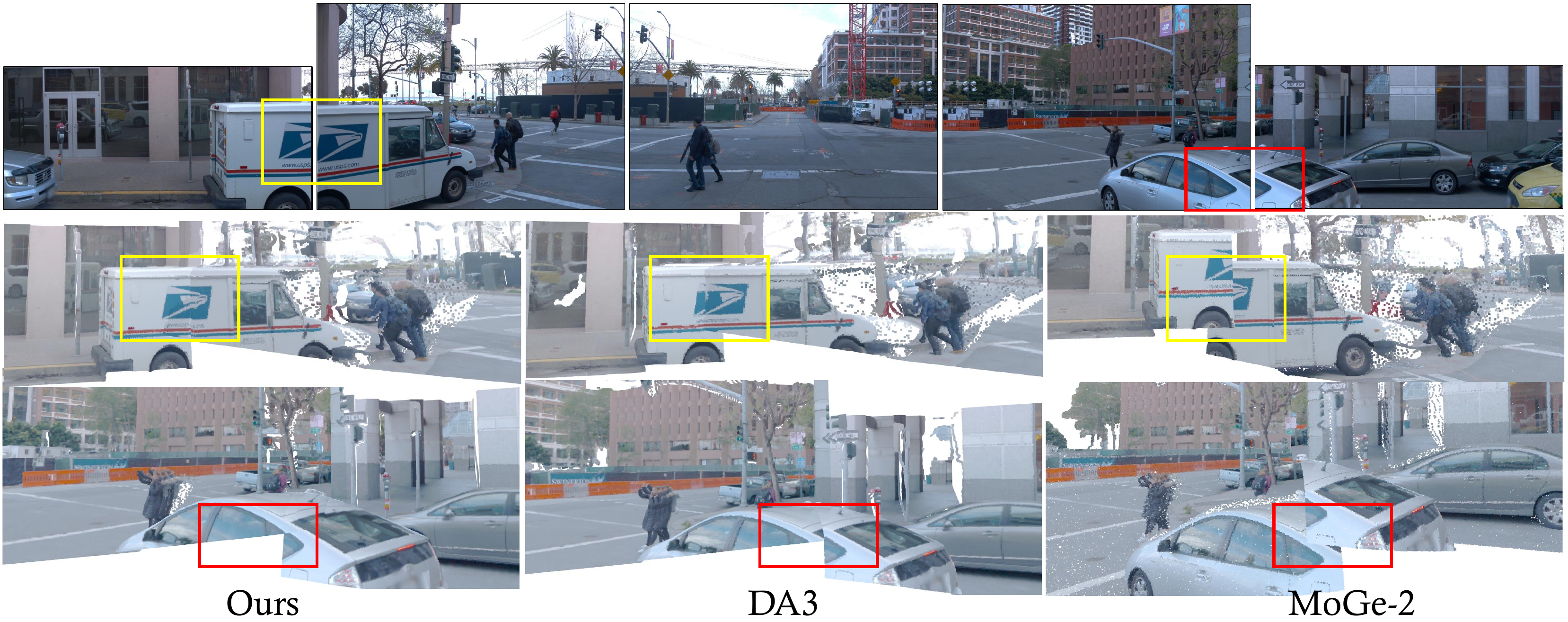}
    \caption{
\textbf{Surround-view visualization on Waymo.}
While DA3\cite{depthanythingv3} shows spatial misalignment under weak low-overlap reasoning, and MoGe-2\cite{mogev22025} suffers from per-view inconsistency, SurroundNEXO produces globally consistent geometry across surrounding scene. 
}
    \label{fig:vis}
\end{figure*}

\textbf{Baselines.}
For metric-scale depth estimation, we organize the compared methods into three categories. First, we include monocular metric depth estimators, including MoGe-2~\cite{mogev22025}, Depth Pro~\cite{DepthPro2024}, UniDepth v2~\cite{unidepthv22025}, and Metric Anything~\cite{metricanything2026} (denoted as MetricAny.). These methods predict metric depth from a single RGB image and serve as strong single-view metric baselines.

Second, we compare with prompt-based metric depth methods, including PromptDA~\cite{2024promptda}, PriorDA~\cite{priorda2025}, Omni-DC~\cite{2025omnicdc}, Lingbot-Depth~\cite{lingbot-depth2026} (denoted as LB-Depth), and Any2Full~\cite{zhou2026any2full}. These methods use sparse depth, prompts, or other external priors to recover metric-scale depth, making them directly relevant to our depth prompting and robustness evaluation.

Third, we evaluate feed-forward 3D foundation models, including VGGT~\cite{2025vggt}, Pi3~\cite{2025pi3}, DA3-Giant~\cite{depthanythingv3}, DVGT~\cite{2025dvgt}, MapAnything~\cite{2026mapanything} (denoted as MapAny.), and OmniVGGT~\cite{omnivggt2025}. These methods can process multiple views or recover 3D structure in a feed-forward manner, and therefore provide important comparisons for both metric depth estimation and 3D reconstruction. For reconstruction experiments, we primarily compare against this feed-forward family.

\begin{table}[t]
    \centering
    \caption{
        \textbf{Single-view metric depth estimation.} \colorbox{first}{\textbf{$1^{st}$}}, \colorbox{second}{$2^{nd}$}, and \colorbox{third}{$3^{rd}$} are highlighted. $^\dagger$ denotes models fine-tuned on the same training datasets.
    }
    \vspace{-0.5em}
    \resizebox{1.0\columnwidth}!{
    \begin{tabular}{lcccccccccccccc}
        \toprule[0.17em]
        {\multirow{2}{*}{\textbf{Method}}} &
        {\multirow{2}{*}{\textbf{Inject}}} &
        {\multirow{2}{*}{\textbf{Alignment}}} &
        \multicolumn{2}{c}{\textbf{Waymo}} &
        \multicolumn{2}{c}{\textbf{NuScenes}} &
        \multicolumn{2}{c}{\textbf{KITTI}} &
        \multicolumn{2}{c}{\textbf{DDAD}} &
        \multicolumn{2}{c}{\textbf{OpenScene}}\\
        \cmidrule(r){4-5} \cmidrule(r){6-7} \cmidrule(r){8-9} \cmidrule(r){10-11} \cmidrule(r){12-13}
        & & &
        Abs Rel $\downarrow$ & $\delta<1.25$ $\uparrow$&
        Abs Rel $\downarrow$ & $\delta<1.25$ $\uparrow$&
        Abs Rel $\downarrow$ & $\delta<1.25$ $\uparrow$&
        Abs Rel $\downarrow$ & $\delta<1.25$ $\uparrow$&
        Abs Rel $\downarrow$ & $\delta<1.25$ $\uparrow$\\
        \midrule
        \textit{Single-view}\\
        Depth-Pro & \xmark & \textcolor{mygreen}{Metric} & 0.390 & 0.230 & 0.236 & 0.522 & 0.135 & 0.876 & 0.450 & 0.176 & 0.259 & 0.477\\
        UniDepth-2 & \xmark & \textcolor{mygreen}{Metric} & 0.147 & 0.829 & 0.120 & \cellcolor{second}0.934 & 0.071 & \cellcolor{third}0.962 & \cellcolor{third}0.091 & \cellcolor{third}0.941 & 0.142 & 0.913\\
        MoGe-2 & \xmark & \textcolor{mygreen}{Metric} & 0.177 & 0.630 & 0.129 & 0.851 & 0.240 & 0.273 & 0.124 & 0.796 & \cellcolor{second}0.094 & \cellcolor{second}0.937\\
        MetricAny. & \cmark & \textcolor{mygreen}{Metric} & 0.097 & \cellcolor{second}0.943 & 0.160 & 0.851 & 0.151 & 0.710 & \cellcolor{second}0.085 & \cellcolor{second}0.944 & 0.152 & 0.860 \\
        \midrule
        \textit{Multi-view}\\
        VGGT & \xmark &\textcolor{magenta}{Scale} & 0.134 & 0.857 & 0.213 & 0.773 & 0.092 & 0.941 & 0.185 & 0.707 & 0.178 & 0.808 \\
        Pi3 & \xmark & \textcolor{magenta}{Scale} & \cellcolor{third}0.096 & 0.914 & 0.122 & 0.869 & \cellcolor{first}\textbf{0.056} & \cellcolor{first}\textbf{0.972} & 0.139 & 0.802 & 0.128 & 0.889\\
        OmniVGGT & \cmark & \textcolor{magenta}{Scale} & 0.104 & 0.925 & 0.194 & 0.762 & 0.095 & 0.927 & 0.195 & 0.682 & 0.267 & 0.814 \\
        MapAny. & \cmark & \textcolor{mygreen}{Metric} & \cellcolor{third}0.096 & 0.911 & \cellcolor{third}0.115 & 0.879 & 0.096 & 0.922 & 0.114 & 0.867 & 0.115 & 0.870 \\
        DVGT & \xmark & \textcolor{mygreen}{Metric} & 0.227 & 0.615 & 0.127 & 0.871 & \cellcolor{third}0.068 & \cellcolor{second}0.969 & 0.097 & 0.922 & 0.110 & 0.909\\
        DA3-G & \cmark & \textcolor{magenta}{Scale} & 0.172 & 0.789 & 0.147 & 0.827 & \cellcolor{second}0.063 & \cellcolor{second}0.969 & 0.142 & 0.812 & 0.132 & 0.875 \\
        DA3-G$^\dagger$ & \cmark & \textcolor{mygreen}{Metric} & \cellcolor{second}0.082 & \cellcolor{third}0.937 & \cellcolor{second}0.112 & \cellcolor{third}0.911 & 0.102 & 0.929 & 0.114 & 0.906 & \cellcolor{third}0.105 & \cellcolor{third}0.916 \\
        \midrule
        \textbf{Ours} & \cmark & \textcolor{mygreen}{Metric} & \cellcolor{first}\textbf{0.048} & \cellcolor{first}\textbf{0.975} & \cellcolor{first}\textbf{0.079} & \cellcolor{first}\textbf{0.950} & 0.070 & \cellcolor{second}0.969 & \cellcolor{first}\textbf{0.077} & \cellcolor{first}\textbf{0.956} & \cellcolor{first}\textbf{0.084} & \cellcolor{first}\textbf{0.938} \\
        \bottomrule[0.17em]
    \end{tabular}
    }
    \label{tab:monodepth}
\end{table}
\begin{table}[t]
    \centering
    \caption{
        \textbf{Cross-view depth consistency evaluation} on NuScenes\cite{nuscenes} and DDAD\cite{ddad}. 
    }
    \resizebox{0.75\columnwidth}!{
    \begin{tabular}{lcccccccccccccc}
        \toprule[0.17em]
        {\multirow{2}{*}{\textbf{Method}}} &
        \multicolumn{4}{c}{\textbf{NuScenes}} &
        \multicolumn{4}{c}{\textbf{DDAD}}\\
        \cmidrule(r){2-5} \cmidrule(r){6-9}
        &
        Abs Rel $\downarrow$ & Sq Rel $\downarrow$ & RMSE $\downarrow$ & $\delta<1.25$ $\uparrow$&
        Abs Rel $\downarrow$ & Sq Rel $\downarrow$ & RMSE $\downarrow$ & $\delta<1.25$ $\uparrow$\\
        \midrule
        \textit{Single-view}\\
        Depth-Pro & 0.353 & 3.482 & 6.461 & 0.476 & 0.625 & 6.654 & 7.772 & 0.276 \\
        UniDepth-2 & \cellcolor{third}0.198 & 2.691 & 6.216 & 0.341 & \cellcolor{second}0.196 & 2.228 & 6.205 & 0.763 \\
        MoGe-2 & 0.235 & 2.683 & 6.281 & 0.680 & 0.347 & 3.598 & 7.274 & 0.544 \\
        MetricAny. & 0.208 & 2.512 & 6.156 & 0.728 & \cellcolor{third}0.198 & 2.340 & 5.834 & \cellcolor{second}0.799 \\
        \midrule
        \textit{Multi-view}\\
        VGGT & 0.220 & 2.272 & 5.608 & 0.660 & 0.303 & 2.522 & 5.680 & 0.629 \\
        Pi3 & 0.211 & 2.387 & 5.959 & 0.707 & 0.321 & 3.571 & 6.865 & 0.604\\
        OmniVGGT & 0.210 & \cellcolor{third}1.977 & \cellcolor{second}4.671 & 0.660 & 0.269 & \cellcolor{second}2.083 & \cellcolor{second}5.336 & 0.562 \\
        MapAny. & 0.328 & 5.348 & 7.212 & \cellcolor{third}0.741 & 0.490 & 8.889 & 8.428 & 0.682 \\
        DVGT & 0.229 & 2.629 & 5.949 & 0.705 & 0.241 & 2.275 & 5.994 & 0.712\\
        DA3-G & 0.217 & 2.111 & 5.443 & 0.684 & 0.314 & 3.633 & 5.945 & 0.576\\
        DA3-G$^\dagger$ & \cellcolor{second}0.173 & \cellcolor{second}1.897 & \cellcolor{third}5.132 & \cellcolor{second}0.801 & 0.238 & \cellcolor{third}2.218 & \cellcolor{third}5.589 & \cellcolor{third}0.769\\
        \midrule
        \textbf{Ours} & \cellcolor{first}\textbf{0.139} & \cellcolor{first}\textbf{1.654} & \cellcolor{first}\textbf{4.527} & \cellcolor{first}\textbf{0.850} & \cellcolor{first}\textbf{0.162} & \cellcolor{first}\textbf{1.840} & \cellcolor{first}\textbf{5.333} & \cellcolor{first}\textbf{0.837}\\
        \bottomrule[0.17em]
    \end{tabular}
    }
    \label{tab:crossview}
\end{table}

\subsection{Depth Estimation}
\textbf{Single-view Metric Depth.}
We first evaluate metric depth from a single view using Abs Rel and $\delta<1.25$. 
While monocular depth estimation models are capable of directly inferring metric depth, multi-view feed-forward models often yield depth maps up to a relative scale. 
Therefore, for models outputting relative depth, we follow standard evaluation protocol by first scaling the predictions using ground-truth depth to obtain metric information prior to computing the metrics.

The results in Tab.~\ref{tab:monodepth} demonstrate that monocular depth estimation (MDE) models can yield depth accuracy highly competitive with that of multi-view feed-forward (FFW) pipelines. 
More importantly, our method establishes new state-of-the-art performance, surpassing existing single- and multi-view baselines across comprehensive metrics on multiple datasets. 
These findings validate the effectiveness of our approach in delivering high-fidelity, metric-scale single-view perception in challenging low-overlap scenes typical of autonomous driving. Notably, feed-forward methods such as Pi3\cite{2025pi3} and DA3\cite{depthanythingv3} exhibit significant advantages on the KITTI dataset. 
We attribute this to KITTI’s specific camera setup, where the left and right cameras have an almost 100\% FoV overlap.
This extensive overlap is particularly well aligned with the methods that exploit dense correspondences. 

\textbf{Cross-view Depth Consistency.}
To evaluate cross-view consistency, we select the nuScenes\cite{nuscenes} and DDAD\cite{ddad} datasets, where the average overlap between their adjacent cameras falls within a limited range of 15\% to 25\%.
Specifically, we evaluate cross-view consistency by reprojecting predicted depths between adjacent cameras using calibrated intrinsics and extrinsics. 
For a given camera pair, the source depth is transformed into the target frame and inverse-warped onto the target image plane. 
We then compute the discrepancy between this warped depth and the target's predicted depth, strictly over valid pixels within their overlapping field of view (FoV). 
Finally, standard metrics (Abs Rel, Sq Rel, RMSE, and $\delta < 1.25$) are computed bidirectionally for all adjacent pairs and averaged to yield the final score.

\begin{figure*}[b!]
    \centering
    \includegraphics[width=\textwidth]{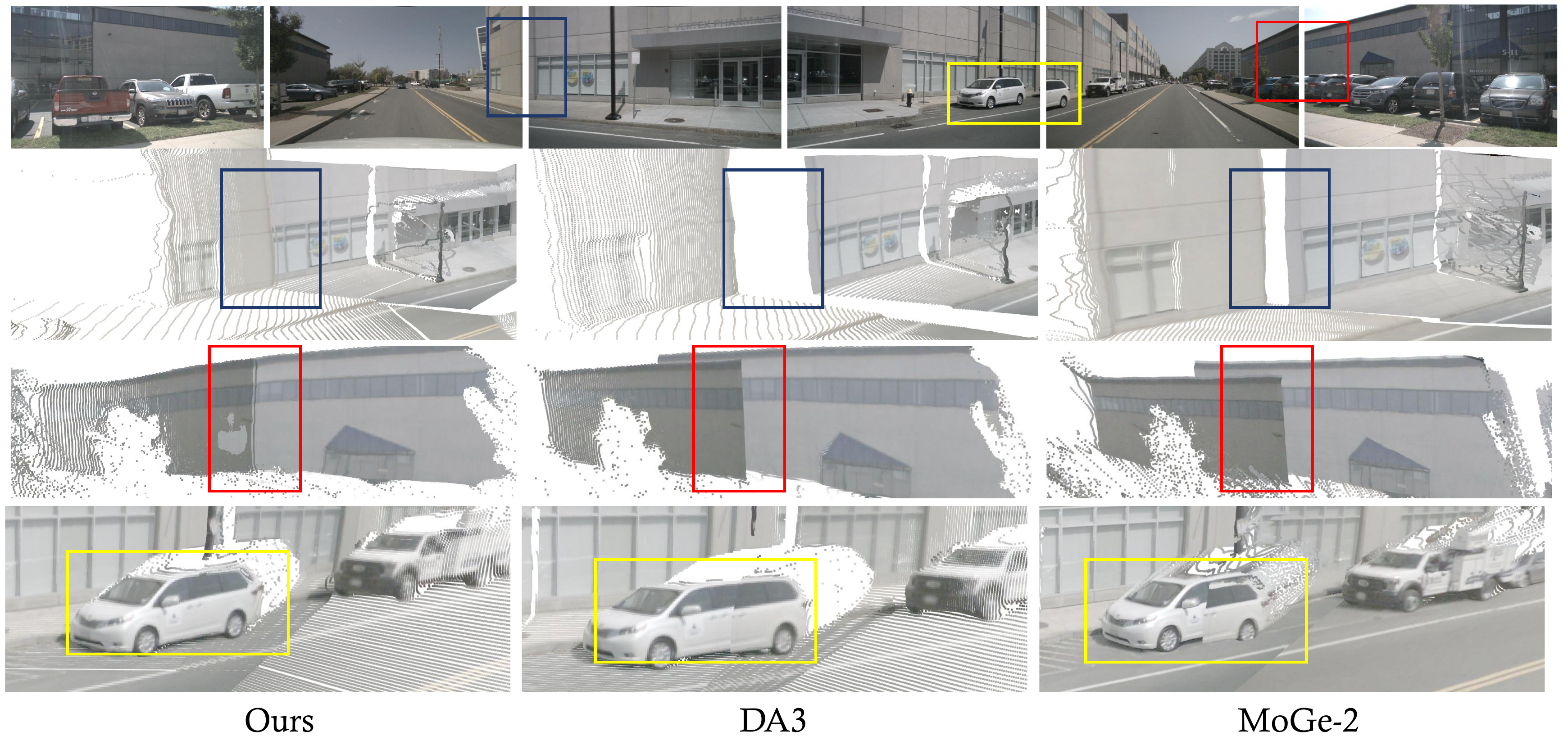}
    \caption{
Surround-view visualization on NuScenes\cite{nuscenes}.
}
    \label{fig:vis2}
\end{figure*}

\begin{figure*}[t]
    \centering
    \includegraphics[width=\textwidth]{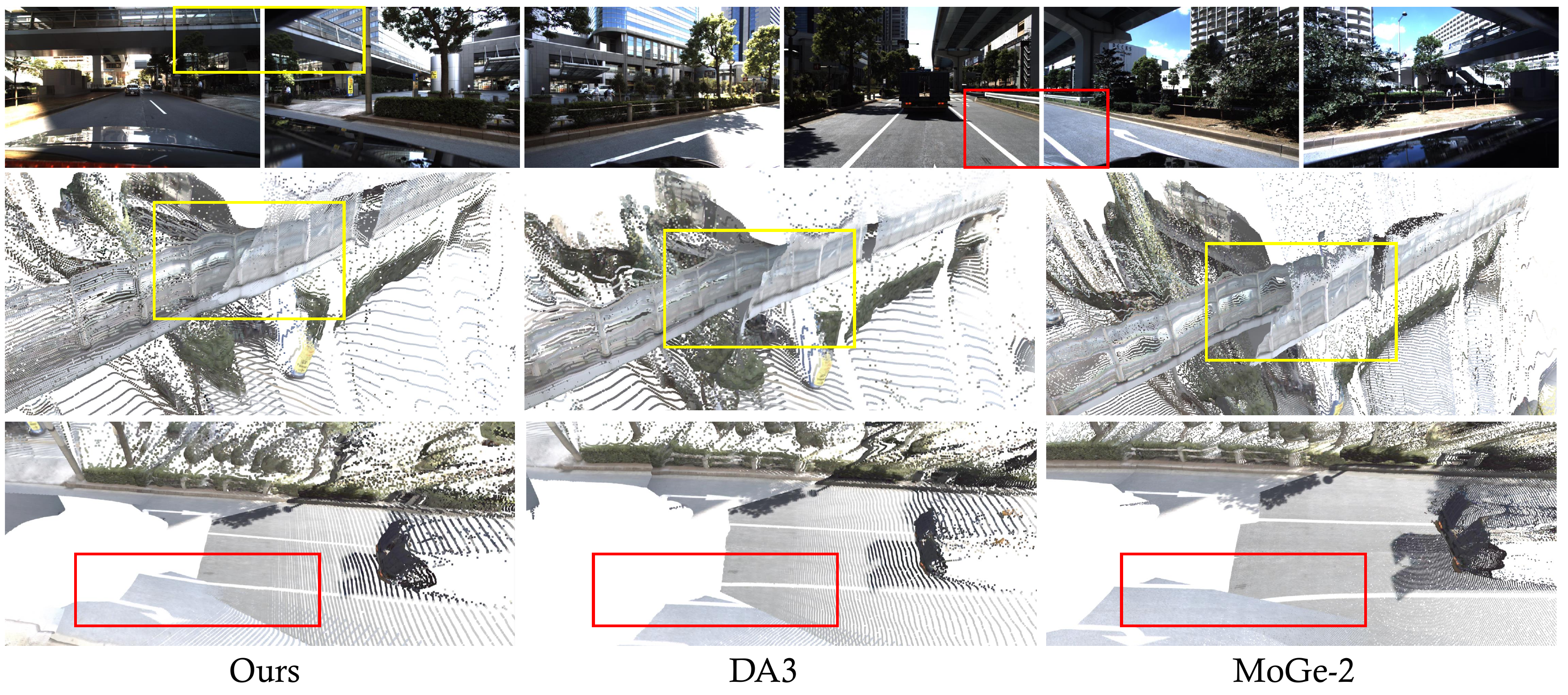}
    \caption{
Surround-view visualization on DDAD\cite{ddad}.
}
    \label{fig:vis3}
    \vspace{-2em}
\end{figure*}

As demonstrated in Tab.~\ref{tab:crossview}, our method achieves the highest cross-view consistency across multiple metrics when compared to current SOTA models. 
We attribute this superior performance to the design of the ERPE module, which provides a shared ray-aligned coordinates for view consistency, thereby significantly enhancing the model's perception of 3D geometric awareness. Fig.~\ref{fig:vis}, Fig.~\ref{fig:vis2} and Fig.~\ref{fig:vis3} further illustrate the qualitative results in Waymo~\cite{waymo}, NuScenes~\cite{nuscenes} and DDAD~\cite{ddad}, which is evident that both MDE and FFW models suffer from severe spatial inconsistencies during the geometric reconstruction of surround-view scenes. 
In contrast, our method maintains superior cross-view consistent geometry, accurately recovering structures across both near-field regions and distant buildings.

\textbf{Prompt Depth Estimation.}
To comprehensively evaluate robustness against diverse depth prompt patterns, we use simulated 4-beam and 8-beam LiDAR scans as well as randomly sampled points at densities of 0.1\% and 1\% as input prompts.
As detailed in Tab.~\ref{tab:promptdepth}, we report Lidar-4, Random-0.1\%, and the overall average across all configurations.
Our approach achieves strong accuracy across conditions and remains robust under extremely sparse prompts, while Lingbot-Depth\cite{lingbot-depth2026} suffers nearly 2$\times$ degradation on Random-0.1\% relative to its overall average.
However, because our approach directly outputs metric depth without relying on GT-based post-alignment, it is evaluated under a more rigorous setting. 
Despite this, our method still matches or surpasses these baselines across the majority of metrics, though it falls slightly short of the optimal on a few metrics.

\begin{table}[t]
    \centering
    \caption{
        \textbf{Prompt depth estimation} on Waymo\cite{waymo} and NuScenes\cite{nuscenes} with various prompt patterns. 
    }
    \resizebox{1.0\columnwidth}!{
    \begin{tabular}{lccccccccccccccccccccc}
        \toprule[0.17em]
        {\multirow{3}{*}{\textbf{Method}}} &
        {\multirow{3}{*}{\textbf{\makecell{Post- \\ alignment}}}} &
        \multicolumn{6}{c}{\textbf{Waymo}} &
        \multicolumn{6}{c}{\textbf{NuScenes}}\\
        \cmidrule(r){3-8} \cmidrule(r){9-14}
        & &
        \multicolumn{2}{c}{Avg.}&
        \multicolumn{2}{c}{Lidar-4}&
        \multicolumn{2}{c}{Random-0.1\%}&
        \multicolumn{2}{c}{Avg.}&
        \multicolumn{2}{c}{Lidar-4}&
        \multicolumn{2}{c}{Random-0.1\%}\\
        \cmidrule(r){3-4}\cmidrule(r){5-6}\cmidrule(r){7-8}\cmidrule(r){9-10}\cmidrule(r){11-12}\cmidrule(r){13-14}
        & &
        Abs Rel $\downarrow$ & RMSE $\downarrow$&
        Abs Rel $\downarrow$ & RMSE $\downarrow$&
        Abs Rel $\downarrow$ & RMSE $\downarrow$&
        Abs Rel $\downarrow$ & RMSE $\downarrow$&
        Abs Rel $\downarrow$ & RMSE $\downarrow$&
        Abs Rel $\downarrow$ & RMSE $\downarrow$\\
        \midrule
        Omni-DC & \xmark & 0.133 & 4.298 & 0.211 & 6.534 & \cellcolor{third}0.208 & \cellcolor{third}5.592 & 0.287 & 6.555 & 0.158 & 4.163 & 0.754 & 15.056 \\
        PromptDA & \xmark & 0.495 & 14.271 & 0.503 & 14.334 & 0.453 & 12.785 & 0.507& 12.047 & 0.507 & 11.717 & 0.538 & 13.331 \\
        PriorDA & \cmark & \cellcolor{second}0.054 & \cellcolor{second}2.258 & \cellcolor{second}0.057 & \cellcolor{second}2.455 & \cellcolor{second}0.053 & \cellcolor{second}2.336 & \cellcolor{first}\textbf{0.084} & \cellcolor{second}2.567 & \cellcolor{first}\textbf{0.075} & \cellcolor{second}2.280 & \cellcolor{second}0.139 & \cellcolor{second}3.976\\
        LB-Depth & \xmark & 0.126 & 4.433 & \cellcolor{third}0.069 & 3.544 & 0.314 & 7.780 & \cellcolor{third}0.173 & \cellcolor{third}5.280 & 0.119 & 4.937 & \cellcolor{third}0.327 & \cellcolor{third}6.560 \\
        Any2Full & \cmark & \cellcolor{third}0.090 & \cellcolor{third}2.850 & 0.117 & \cellcolor{third}2.484 & 0.325 & 6.719 & 0.542 & 6.946 & \cellcolor{second}0.080 & \cellcolor{third}2.754 & 1.927 & 19.469\\
        \textbf{Ours} & \xmark & \cellcolor{first}\textbf{0.053} & \cellcolor{first}\textbf{1.946} & \cellcolor{first}\textbf{0.054} & \cellcolor{first}\textbf{1.989} & \cellcolor{first}\textbf{0.049} & \cellcolor{first}\textbf{1.843} & \cellcolor{second}0.087 & \cellcolor{first}\textbf{2.005} & \cellcolor{third}0.083 &
        \cellcolor{first}\textbf{1.893} &\cellcolor{first}\textbf{0.103} &\cellcolor{first}\textbf{2.343}\\
        \bottomrule[0.17em]
    \end{tabular}
    }
    \label{tab:promptdepth}
\end{table}
\begin{table}[t]
    \centering
    \caption{
        \textbf{Quantitative results of 3D reconstruction.}  \colorbox{first}{\textbf{$1^{st}$}}, \colorbox{second}{$2^{nd}$}, and \colorbox{third}{$3^{rd}$} are highlighted.
    }
    \vspace{-0.5em}
    \resizebox{1.0\columnwidth}!{
    \begin{tabular}{lcccccccccccccc}
        \toprule[0.17em]
        {\multirow{2}{*}{\textbf{Method}}} &
        {\multirow{2}{*}{\textbf{Inject}}} &
        {\multirow{2}{*}{\textbf{Alignment}}} &
        \multicolumn{2}{c}{\textbf{Waymo}} &
        \multicolumn{2}{c}{\textbf{NuScenes}} &
        \multicolumn{2}{c}{\textbf{KITTI}} &
        \multicolumn{2}{c}{\textbf{DDAD}} &
        \multicolumn{2}{c}{\textbf{OpenScene}}& \textbf{Time}\\
        \cmidrule(r){4-5} \cmidrule(r){6-7} \cmidrule(r){8-9} \cmidrule(r){10-11} \cmidrule(r){12-13}
        & & &
        Acc. $\downarrow$& Comp. $\downarrow$&
        Acc. $\downarrow$& Comp. $\downarrow$&
        Acc. $\downarrow$& Comp. $\downarrow$&
        Acc. $\downarrow$& Comp. $\downarrow$&
        Acc. $\downarrow$& Comp. $\downarrow$\\
        \midrule
        VGGT & \xmark &\textcolor{magenta}{Scale} & 0.554 & 0.290 & 0.603 & 0.346 & 0.182 & 0.258 & 0.410 & 0.272 & 0.246 & 0.219 & \textasciitilde55.62s\\
        Pi3 & \xmark & \textcolor{magenta}{Scale} & 0.585 & 0.258 & 0.400 & \cellcolor{third}0.249  & \cellcolor{second}0.144 & \cellcolor{second}0.111 & 0.310 & \cellcolor{third}0.172 & 0.249 & \cellcolor{second}0.171 & \textasciitilde35.08s\\
        OmniVGGT & \cmark & \textcolor{magenta}{Scale} & 0.525 & 0.387 & 0.526 & 0.430 & 0.176 & 0.326 & 0.409 & 0.513 & 0.268 & 0.432 & \textasciitilde45.59s\\
        DVGT & \xmark & \textcolor{mygreen}{Metric} & 0.756 & 0.481 & 1.078 & 0.808 & 0.522 & 0.743 &  0.371 & 0.705 & 0.306 & 0.381 & \cellcolor{first}\textasciitilde19.73s\\
        DA3-G & \cmark & \textcolor{magenta}{Scale} & \cellcolor{third}0.333 & \cellcolor{third}0.216 & \cellcolor{third}0.360 & 0.275 & \cellcolor{third}0.155 & 0.128 & \cellcolor{second}0.252 & 0.212 & \cellcolor{third}0.206 & 0.222 & \cellcolor{third}\textasciitilde31.79s \\
        DA3-G$^\dagger$ & \cmark & \textcolor{mygreen}{Metric} & 0.335 & \cellcolor{second}0.210 & 0.381 & 0.256 & 0.222 & 0.156 & \cellcolor{third}0.295 & 0.195 & 0.270 & 0.244 &\cellcolor{third}\textasciitilde31.79s\\
        MapAny. & \cmark & \textcolor{mygreen}{Metric} & \cellcolor{second}0.288 & 0.361  & \cellcolor{second}0.330 & \cellcolor{second}0.201 & \cellcolor{first}\textbf{0.125} & \cellcolor{first}\textbf{0.100} & 0.361 & \cellcolor{second}0.150 & \cellcolor{second}0.198 & \cellcolor{third}0.190 & \textasciitilde34.49s\\
            \textbf{Ours} & \cmark & \textcolor{mygreen}{Metric} & \cellcolor{first}\textbf{0.223} & \cellcolor{first}\textbf{0.133} & \cellcolor{first}\textbf{0.281} & \cellcolor{first}\textbf{0.173} & 0.170 & \cellcolor{third}0.125 & \cellcolor{first}\textbf{0.221} & \cellcolor{first}\textbf{0.128} & \cellcolor{first}\textbf{0.145} & \cellcolor{first}\textbf{0.116} & \cellcolor{second}\textasciitilde25.12s\\
        \bottomrule[0.17em]
    \end{tabular}
    }
    \label{tab:reconstruction}
\end{table}

\subsection{3D Reconstruction}
For the 3D point cloud reconstruction evaluation, valid pixels are first extracted from the sparse LiDAR depth. 
The model's depth predictions at these specific locations are unprojected into 3D space via camera parameters and aligned using the Iterative Closest Point (ICP) algorithm to calculate Accuracy (Acc) and Completeness (Comp). 
For models outputting relative depth, scale alignment is performed beforehand to recover the metric scale.
As shown in Tab.~\ref{tab:reconstruction}, our model achieves accurate, metric-scale point cloud reconstruction on low-overlap datasets. 
Notably, it outperforms MapAnything\cite{2026mapanything}, the SOTA feed-forward model that also supports injection.
We also report inference time for processing $24\times8$ images on a single 96GB NVIDIA H20 GPU.
Under this setting, our method requires only \textasciitilde25.12s, achieving substantially better reconstruction quality than most feed-forward baselines while remaining faster.
Nevertheless, we still observe a slight performance gap on the KITTI dataset, which stems from the identical factors discussed during depth estimation analysis.

\subsection{Zero Shot Generalization}
Furthermore, to rigorously assess zero-shot generalization capabilities, we additionally evaluate our method on the unseen Argoverse\cite{Argoverse2} dataset, reporting metrics for both single-view depth estimation and 3D reconstruction.
As shown in Tab.~\ref{tab:zeroshot}, DVGT\cite{2025dvgt} fails to achieve satisfactory generalization when confronted with unseen camera layouts. 
In contrast, our model delivers superior metric depth accuracy and reconstructed point cloud quality compared to general feed-forward models. 
This demonstrates that our approach successfully maintains robust performance when tackling scene reconstruction under novel configurations.

\begin{table}[h]
    \centering
    \caption{
        \textbf{Zero-shot generalization on Argoverse.}
    }
    \resizebox{0.5\linewidth}{!}{%
    \begin{tabular}{lccccc} 
        \toprule[0.17em]
        \multirow{2}{*}{\textbf{Method}} &
        \multirow{2}{*}{\textbf{Alignment}} &
        \multicolumn{4}{c}{\textbf{Argoverse}} \\
        \cmidrule(r){3-6}
        & &
        Abs Rel $\downarrow$ & $\delta<1.25$ $\uparrow$&
        Acc. $\downarrow$& Comp. $\downarrow$\\
        \midrule
        VGGT &\textcolor{magenta}{Scale} &  0.134 & \cellcolor{third}0.870 & 0.686 & 0.251 \\
        Pi3 & \textcolor{magenta}{Scale} & \cellcolor{second}0.101 & \cellcolor{second}0.921 & 0.943 & \cellcolor{third}0.227  \\
        OmniVGGT& \textcolor{magenta}{Scale} & 0.205 & 0.758 & 0.629 & 0.654\\
        DVGT& \textcolor{mygreen}{Metric} & 0.491 & 0.387 & 3.383 & 0.881 \\
        DA3-G & \textcolor{magenta}{Scale} & 0.236 & 0.732 & \cellcolor{second}0.353 & 0.243\\
        MapAny. & \textcolor{mygreen}{Metric} & \cellcolor{third}0.113 & 0.863 & \cellcolor{third}0.478 & \cellcolor{second}0.165\\
        \textbf{Ours} & \textcolor{mygreen}{Metric} & \cellcolor{first}\textbf{0.081} & \cellcolor{first}\textbf{0.954} & \cellcolor{first}\textbf{0.337} & \cellcolor{first}\textbf{0.160}\\
        \bottomrule[0.17em]
    \end{tabular}%
    }
    \label{tab:zeroshot}
\end{table}

\subsection{Ablation Study}
For efficient comparison, all ablation models are initialized from the pretrained DA3-L\cite{depthanythingv3} model and trained under the same setting within each ablation group.

\begin{wraptable}{r}{0.5\textwidth}
    \centering
    \caption{
        \textbf{Ablation on proposed modules.} 
    }
    \resizebox{\linewidth}{!}{%
    \begin{tabular}{lcccccccccc} 
        \toprule[0.17em]
         & & & & \multicolumn{2}{c}{\textbf{Single-view}} & \multicolumn{2}{c}{\textbf{Cross-view}}\\
        \cmidrule(r){5-6} \cmidrule(r){7-8} 
        {\textbf{\#}} &
        {\textbf{SMA}} &
        {\textbf{ERPE}} &
        {\textbf{Stage}} &
        Abs Rel $\downarrow$ & $\delta<1.25$ $\uparrow$&
        Abs Rel $\downarrow$ & $\delta<1.25$  $\uparrow$ \\
        \midrule
        1 &  & & all global & 0.236 & 0.627 & 0.261 & 0.710\\
        2 & \cmark & & all global & 0.120 & 0.893 & 0.197 & 0.749\\
        3 & \cmark & & global first & 0.114 & 0.905 & 0.188 & 0.760\\
        4 & \cmark & & progressive & 0.106 & 0.915 & 0.185 & 0.763\\
        5 & \cmark & \cmark & progressive & \textbf{0.093} & \textbf{0.928} & \textbf{0.151} & \textbf{0.835}
        \\
        \bottomrule[0.17em]
    \end{tabular}%
    }
    \vspace{-1em}
    \label{tab:ablation1}
\end{wraptable}

\textbf{Component-wise Ablation.}
To investigate the contribution of each component, we conduct evaluation on NuScenes\cite{nuscenes}.
In Tab.~\ref{tab:ablation1}, adding SMA to the all-global baseline substantially reduces single-view error, confirming that sparse metric anchors provide essential metric scale cues.
Among attention-stage designs, the proposed progressive strategy outperforms both all-global and global-first variants, indicating that gradually expanding token interaction is more effective than introducing global attention prematurely.
Finally, ERPE further improves the progressive model, especially for cross-view consistency.
These results verify that SMA, ERPE and PGT respectively contribute metric grounding, globally consistent cross-view geometry and robust interaction scheduling.

\begin{wraptable}{r}{0.5\textwidth}
    \centering
    \caption{
        \textbf{Ablation on inject method.} 
    }
    \resizebox{\linewidth}{!}{%
    \begin{tabular}{lccccc} 
        \toprule[0.17em]
        {\textbf{Method}} & 
        Abs Rel $\downarrow$ & Sq Rel $\downarrow$ & RMSE $\downarrow$ & $\delta<1.25$ $\uparrow$\\
        \midrule
        DPT Head & 0.105 & 1.962 & 1.340 & 0.918 \\
        Patch Embed & 0.118 & 2.187 & 1.218 & 0.904\\
        Scale Head & 0.112 & 2.257 & 1.209 & 0.905\\
        \textbf{Ours} & \textbf{0.086} & \textbf{1.945} & \textbf{0.919} & \textbf{0.947}\\
        \bottomrule[0.17em]
    \end{tabular}%
    }
    \vspace{-1em}
    \label{tab:ablation2}
\end{wraptable}

\textbf{Depth-injection Ablation.}
We further evaluate how depth information should be injected for metric-scale recovery on NuScenes\cite{nuscenes}.
As reported in Tab.~\ref{tab:ablation2}, directly injecting depth into the DPT head, concatenating depth patch embeddings with image tokens, or predicting a global scale through an additional scale head all provide less accurate metric estimates than SMA.
This suggests that sparse depth should not be treated merely as an output-side correction or an additional input modality.
Instead, SMA enables sparse metric observations to interact with dense visual tokens inside the network, leading to more accurate and robust metric-scale depth prediction.
\section{Conclusion}
We presented \textbf{SurroundNEXO}, a metric-grounded surround-view depth framework for low-overlap autonomous driving scenes. 
Rather than searching for dense visual correspondences, SurroundNEXO establishes a shared ego-centric metric space in which weakly overlapping views can exchange geometry in a controlled and scale-aware manner.
This design enables depth prediction that is both metrically accurate and spatially consistent across weakly overlapping camera views. 
Extensive experiments across depth estimation, cross-view consistency, sparse-prompt robustness, zero-shot transfer, and 3D reconstruction demonstrate the effectiveness of our approach on diverse driving benchmarks. 
Overall, SurroundNEXO provides a practical geometry foundation for calibration-aware surround-view perception and downstream scene reconstruction in autonomous driving.

\clearpage
\nocite{*}
\bibliographystyle{plainnat}
\bibliography{main}

\end{document}